\documentclass[letterpaper, 10 pt, conference]{ieeeconf}    

\IEEEoverridecommandlockouts    

\usepackage{graphics} 
\usepackage{times} 
\usepackage{amsmath} 
\usepackage{amssymb}  
\usepackage[bookmarks=false,hidelinks]{hyperref}
\let\labelindent\relax
\usepackage{enumitem}
\usepackage{float}
\usepackage{multirow}
\usepackage{booktabs}
\usepackage{tabularray}
\usepackage{array}
\usepackage{colortbl}
\usepackage{adjustbox}
\usepackage{balance}

\usepackage{orcidlink}

\usepackage{ulem}

\newcommand{\revise}[1]{{#1}}

\title{PhysMirror: Physics-Aware Mirror Object Generation}

\author{Xuan-Bach Mai$^{1,\dagger}$\orcidlink{0009-0007-0038-5261}, Duy-Phuc Nguyen$^{1,\dagger}$\orcidlink{0009-0008-3690-7743}, Quoc-Van Le$^{1}$\orcidlink{0009-0009-5271-229X}, Tam V. Nguyen$^{2, \ddagger}$\orcidlink{0000-0003-0236-7992}, Thanh-Toan Do$^{3}$\orcidlink{0000-0002-6249-0848}, \\ Huu Le$^{4}$, Duong-Van Nguyen$^{4}$\orcidlink{0000-0002-5122-0543}, Minh-Triet Tran$^{1}$\orcidlink{0000-0003-3046-3041}, and Trung-Nghia Le$^{1,\ddagger,*}$\orcidlink{0000-0002-7363-2610}
\thanks{$^{1}$Xuan-Bach Mai, Duy-Phuc Nguyen, Quoc-Van Le, Minh-Triet Tran, and Trung-Nghia Le are with University of Science, Ho Chi Minh City, Vietnam and with Vietnam National University, Ho Chi Minh City, Vietnam {\tt\small \{mxbach22, ndphuc22, lqvan22\}@apcs.fitus.edu.vn, \{tmtriet, ltnghia\}@fit.hcmsu.edu.vn}}%
\thanks{$^{2}$Tam V. Nguyen is with University of Dayton, Dayton, Ohio, US
        {\tt\small tamnguyen@udayton.edu}}%
\thanks{$^{3}$Thanh-Toan Do is with Monash University, Melbourne, Victoria, Australia
        {\tt\small toan.do@monash.edu}}%
\thanks{$^{4}$Huu Le and Duong-Van Nguyen are with VinFast and with VinUniversity, Ha Noi, Vietnam
        {\tt\small \{huu.lee, duongvan.nguyen\}@vinfastauto.com}}%
\thanks{$^{\dagger}$ These authors have equal contributions}%
\thanks{$^{\ddagger}$ These authors are co-supervisors}%
\thanks{$^{*}$ Corresponding author}%
}

\begin{document}

\maketitle
\thispagestyle{empty}
\pagestyle{empty}

\begin{abstract}
Synthesizing physically accurate mirror reflections remains a fundamental challenge for modern text-to-image diffusion models, which are increasingly critical for generating synthetic training data for embodied AI and robotic perception. These models typically struggle with strict geometric constraints, leading to hallucinations that degrade the utility of the synthetic data. To address this, we introduce a novel, end-to-end physics-aware generation framework namely PhysMirror that natively enforces projective geometry through explicit 3D spatial priors. Our method automatically lifts prompted objects into 3D meshes and constructs a lightweight, mathematically exact mirror scene within a simulated environment. By rendering this explicit 3D scene, we extract precise 2D conditioning elements, such as depth maps and segmentation maps, that serve as robust guiding signals for downstream diffusion models, guiding them to generate images with physically correct mirror reflections. Moreover, we introduce Mirror Consistency Score (MCS), reference-free, fully automated metric that quantifies physical correctness using dense feature matching and vanishing point convergence. Experimental results on our newly constructed MirrOB dataset demonstrate that our approach outperforms state-of-the-art baselines in reflection accuracy and physical realism, while maintaining strong text-to-image semantic alignment, providing a reliable pipeline for embodied AI data generation. \revise{The source code is released at \url{https://duyphuc0701.github.io/PhysMirror}.}


\end{abstract}

\section{Introduction}
\begin{figure}[tp]
    \centering
    \includegraphics[width=\linewidth]{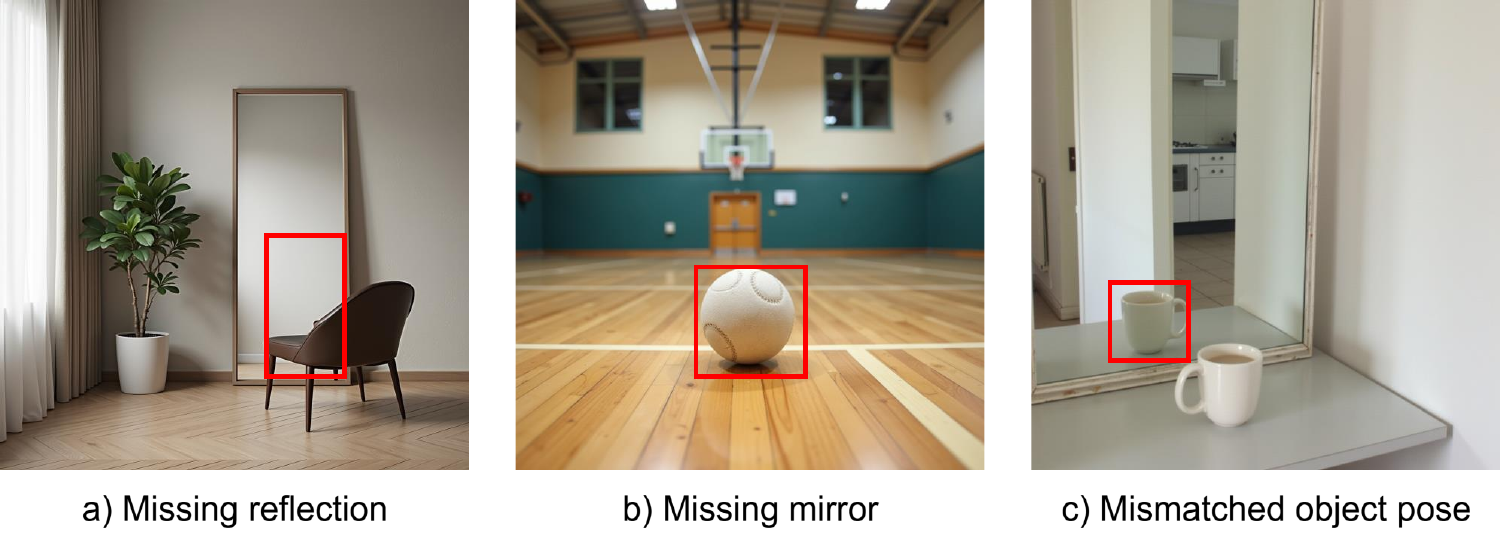}
    \vspace{-5mm}
    \caption{Typical failure cases of the FLUX.1-dev text-to-image model when generating mirror scenes. Common errors include (a) missing the reflection entirely, (b) missing the mirror itself, and (c) generating a mismatched object pose between the real object and its reflection.}
    \label{fig:flux-failure-cases}
    \vspace{-5mm}
\end{figure}

Visual perception in complex, real-world environments poses significant challenges for autonomous agents and robotic systems. Mirrors, in particular, frequently cause severe failures in robot depth estimation, spatial mapping, and manipulation tasks due to illusory depth and reflection mismatches. While synthetic datasets generated by modern text-to-image (T2I) diffusion models, such as FLUX.1~\cite{flux2024} and Stable Diffusion families~\cite{podell2023sdxl, esser2024scaling}, offer an exciting way to train embodied AI, these models often struggle to obey to physical laws and geometric consistency. A critical failure case is the synthesis of physically correct mirror reflections. Generating realistic reflections using these T2I models often results in structural hallucinations or perspective mismatches, compromising the utility of the generated data. As illustrated in Fig.~\ref{fig:flux-failure-cases}, models like FLUX.1-dev~\cite{flux2024} frequently generates obvious errors such as missing reflections, omitted mirrors, or mismatched object poses where the reflection's orientation physically contradicts the real object.

Recent approaches~\cite{dhiman2025reflecting, dhiman2025mirrorverse} attempt to address this by reframing reflection generation as an inpainting task. However, these methods treat reflections as a 2D completion problem rather than a strict geometric projection, failing to explicitly enforce the underlying mathematical symmetry properties. To overcome these limitations and ensure true geometric fidelity for robotic training data, we hypothesize that explicit 3D spatial reasoning must be incorporated. The rapid advancement of text-to-3D models~\cite{jun2023shap, xiang2025structured} and differentiable rendering~\cite{ravi2020accelerating, nimier2019mitsuba} allows us to automatically construct lightweight, modular 3D environments to extract precise spatial conditioning maps. This approach facilitates the rapid generation of physically reliable synthetic data, which can significantly enhance the training of perception networks.

In this work, we introduce PhysMirror, a novel end-to-end physics-aware generation pipeline designed to create photorealistic images with accurate mirror reflections. By utilizing a text-to-3D generative model, we generate the underlying object mesh directly from the input prompt. We then simulate the geometric relationship between the object and a virtual mirror plane within a 3D rendering framework~\cite{ravi2020accelerating}. Through this process, we can automatically render spatial constraints, such as depth and segmentation maps, from optimized viewpoints. These 2D spatial maps can be integrated into diffusion architectures through zero-shot conditioning frameworks~\cite{tan2025ominicontrol, li2025seg2any} or custom adapters, successfully bridging the gap between semantic creativity and physical laws. \revise{The source code is released at \url{https://github.com/T2I-Mirror-Object/PhysMirror}.}

\begin{figure*}[tp]
    \centering
    \includegraphics[width=\textwidth]{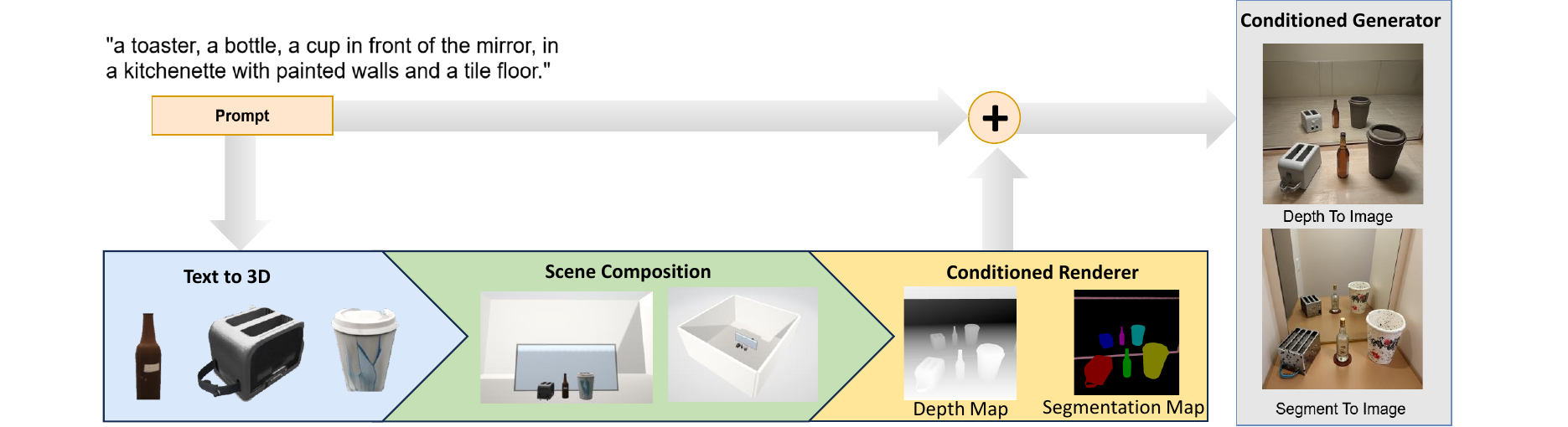}
    \vspace{-5mm}
    \caption{Overview of the proposed PhysMirror. Given a text prompt, primary objects are parsed and lifted into 3D meshes via a text-to-3D model (Stage 1). These meshes are then assembled into a physically accurate 3D mirror scene (Stage 2). Next, a virtual camera is positioned to render precise spatial maps, such as depth and segmentation (Stage 3). Finally, these conditioning maps guide a text-to-image generative model to synthesize the final photorealistic image with geometrically correct mirror reflections (Stage 4).}
    \label{fig:proposed_method}
\end{figure*}

Furthermore, to address the lack of automated evaluation tools in this domain, we introduce Mirror Consistency Score (MCS). By leveraging dense semantic feature matching and the principles of projective geometry, MCS evaluates the physical correctness of generated reflections based on vanishing point convergence. This fully automated metric provides a mathematically rigorous, reference-free measurement, making it highly reliable for benchmarking physics-aware generative models without requiring human intervention. In summary, our contributions are as follows:
\begin{itemize}
    \item We present an end-to-end, fully automatic pipeline that constructs a geometrically consistent 3D scene directly from text prompts, utilizing it to extract explicit geometric priors. To the best of our knowledge, we are the first to address text-to-image generation with physically correct mirror reflections.
    \item We propose the Mirror Consistency Score (MCS), a novel, fully automated evaluation metric that quantitatively measures the physical correctness of mirror reflections without requiring ground-truth annotations or human intervention, by analyzing the projective consistency between objects and their reflected counterparts.
    \item We construct a new Mirror Object Benchmark (MirrOB) dataset, and conduct extensive experiments, demonstrating that our custom depth-conditioned LoRA and zero-shot conditioning frameworks significantly surpass state-of-the-art text-to-image baseline models, such as FLUX.1-dev and SDXL, in geometric realism and mirror reflection accuracy while maintaining strong semantic text-image alignment.
\end{itemize}

\section{Related Work}
\textbf{Image Generation with Diffusion Models.} The field of image synthesis has been revolutionized by denoising diffusion probabilistic models (DDPMs)~\cite{rombach2022high}. Early large-scale models demonstrated unprecedented capabilities in generating high-fidelity images from natural language descriptions. More recent architectures, such as FLUX~\cite{flux2024} and Stable Diffusion 3.5~\cite{esser2024scaling}, have further improved image quality and prompt adherence by scaling up parameters and incorporating advanced transformer-based backbones instead of traditional U-Net backbones. While these foundation models excel at synthesizing realistic textures and lighting, they often struggle with complex spatial relationships and physical constraints~\cite{liu2025generative, meng2024phybench}. To address these controllability issues, several works have introduced adapter-based mechanisms, such as ControlNet~\cite{zhang2023adding} and T2I-Adapter~\cite{mou2024t2i}. These methods allow users to guide the generation process using additional spatial conditions like edge maps, depth maps, or segmentation masks. These approaches typically rely on user-provided conditions or inferred maps from existing images.

\textbf{Text-to-3D Generation.} Generating 3D content directly from text is a rapidly evolving field. Optimization-based methods, such as DreamFusion~\cite{poole2022dreamfusion} and DreamGaussian~\cite{tang2023dreamgaussian}, effectively lift 2D diffusion priors into 3D space via score distillation sampling, though they often suffer from slow convergence and the ``Janus problem'' (multi-faced objects). Feed-forward models such as Shap-E~\cite{jun2023shap} and TRELLIS~\cite{xiang2025structured} offer significantly faster inference by learning a direct mapping from text to explicit 3D meshes.

\textbf{Mirror Reflection Synthesis.} The generation of realistic mirror reflections remains a specific and challenging subset of physics-aware image synthesis. Most works attempt to solve the challenge of reconstructing mirror reflections in a 3D scene for the novel-view synthesis task~\cite{liu2024mirrorgaussian, zeng2023mirror, meng2024mirror}. Notably, previous methods~\cite{dhiman2025reflecting} formulated reflection generation as a depth-conditioned inpainting problem, attempting to fill in the reflection based on the surrounding context. Similarly, large-scale dataset and training curriculum have been proposed to improve reflection realism~\cite{dhiman2025mirrorverse}. In this work, we aim to build an automated pipeline to generate the photorealistic image, with objects placed in front of a mirror, directly from text prompt, without relying on inpainting-based formulations.

\section{Proposed Method}

\begin{figure*}[tp]
    \centering
    \includegraphics[width=0.9\textwidth]{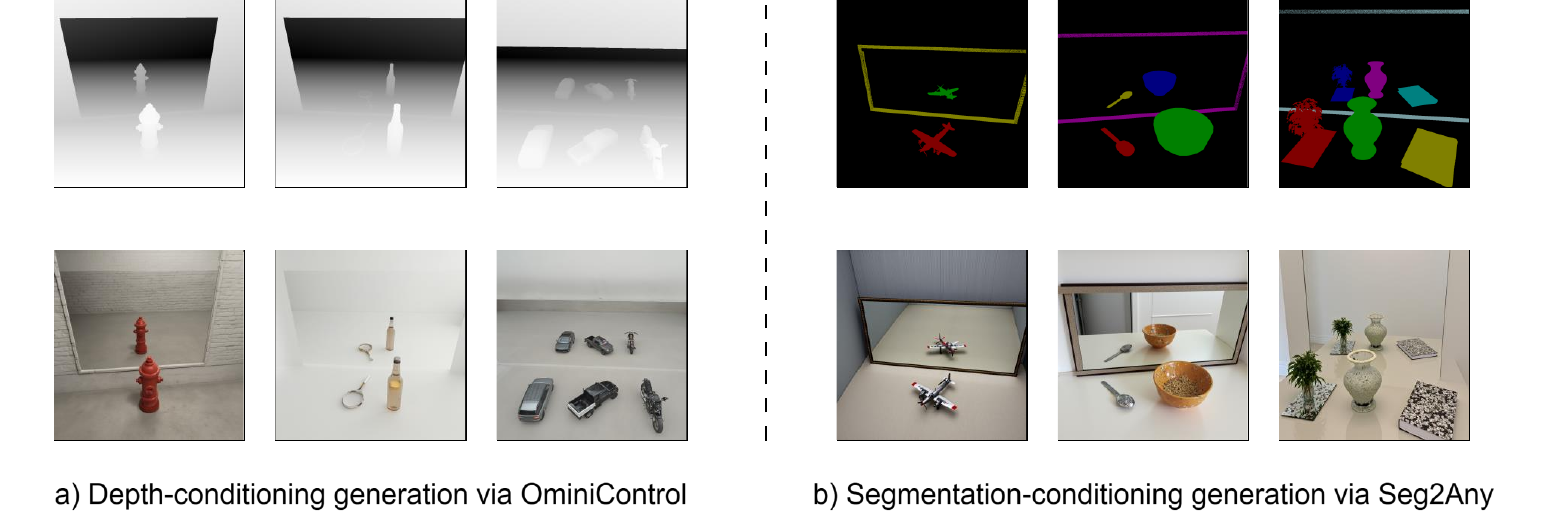}
    \vspace{-3mm}
    \caption{Conditioning-guided generation strategies. (a) The depth map rendered from the 3D scene is passed to a depth-conditioned model (OminiControl~\cite{tan2025ominicontrol}) to generate the final image. (b) The segmentation map, together with per-region text prompts, is passed to Seg2Any~\cite{li2025seg2any} for region-level semantic control.}
    \label{fig:conditioning-strategies}
    \vspace{-3mm}
\end{figure*}

We propose PhysMirror which is a modular pipeline that transitions from a text prompt to a physically grounded 3D scene, then projects back to 2D spatial conditioning maps to guide photorealistic image generation. As illustrated in Fig.~\ref{fig:proposed_method}, the pipeline consists of four stages: (1) lifting prompted objects into 3D meshes, (2) composing a physically accurate mirror scene, (3) rendering conditioning maps from a carefully chosen viewpoint, and (4) guiding a text-to-image generative model with these spatial priors.

\subsection{3D Mesh Generation}
\label{sec:3d_mesh_generation}
The foundation of our physics-aware pipeline relies on establishing an explicit 3D representation for each object within the scene. Given an input text prompt, we parse the text to identify primary objects described in the scene. Each extracted object name is then passed to a text-to-3D generative model to lift the semantic concept into an explicit 3D mesh. While various off-the-shelf text-to-3D models~\cite{poole2022dreamfusion, tang2023dreamgaussian, jun2023shap, xiang2025structured} can be employed, the critical requirement is strong shape fidelity. This geometric quality propagates to all later stages, because accurate object meshes lead to more precise depth and segmentation maps, ultimately yielding more coherent reflections in the final image. 

\subsection{Scene Composition}
\label{sec:scene_composition}
A central contribution of our pipeline is the explicit construction of a complete 3D scene with objects being placed in front of a mirror. Unlike 2D completion approaches~\cite{dhiman2025reflecting, dhiman2025mirrorverse}, computing reflections within a simulated 3D environment is physically exact and geometrically simple. It is a exact planar reflection that can be applied in closed form with no need for approximation or generative inpainting. Furthermore, a single 3D scene can be observed from many viewpoints, which naturally enriches the diversity of generated images.

\textbf{Scene layout.}
We first define a global right-handed world coordinate system ($X, Y, Z$). We establish the origin (0, 0, 0) at the bottom-center of the mirror, exactly where it intersects the floor plane. The $X$-axis extends horizontally across the mirror plane, the $Y$-axis points vertically upward (making the floor the $XZ$-plane where $Y=0$), and the $Z$-axis projects perpendicularly outward from the mirror surface. Consequently, the mirror lies exactly on the $XY$-plane ($Z=0$). The scene consists of the floor plane, a set of room walls, the mirror, the primary objects placed in the physical space in front of the mirror ($Z > 0$). Without a floor, a generative downstream model lacks grounding signal to distinguish objects resting on a surface from objects floating in mid-air. Similarly, without walls, the model lacks context for the mirror's placement, which often leads to spatially inconsistent backgrounds and structural artifacts.

\textbf{Object placement.}
The primary objects placed in front of the mirror. Each object is grounded so that its bottom surface sits at floor level. To increase diversity, each object may undergo a random rotation around its vertical axis, while keeping a small, strict distance from the mirror.

\textbf{Reflection computation.}
Given the mirror as a plane with unit normal $\mathbf{n}$ at signed distance $d$ from the origin, the reflection $\mathbf{p}'$ of any point $\mathbf{p}$ on a 3D object surface visible to the mirror is given by:

\begin{equation}
    \mathbf{p}' = \mathbf{p} - 2\,(\mathbf{n} \cdot \mathbf{p} - d)\,\mathbf{n}.
    \label{eq:reflection}
\end{equation}

Applying Eq.~\eqref{eq:reflection} to every vertex of the object mesh produces a geometrically correct reflected mesh, automatically placed at the optically correct position behind the mirror plane. Because this transformation is applied directly to explicit 3D geometry, the resulting reflection serves as a mathematical ground truth with no generative uncertainty. Additionally, the winding order of the reflected faces is reversed to ensure correct surface normals for rendering.

\subsection{Spatial Conditioning Extraction}
\label{sec:spatial_conditioning_extraction}
With the 3D scene fully assembled, we simulate a virtual camera and use a standard 3D rasterization engine to render the explicit spatial conditioning maps that guide the image generation process.

\textbf{Camera placement.}
The virtual camera is positioned in the physical space at a specific coordinate ($X_c, Y_c, Z_c$) where $Z_c > 0$. A key design choice is that the camera is not placed directly on the mirror's central normal axis (where $X_c = 0$). Instead, the camera's azimuth angle is randomly sampled to introduce a deliberate horizontal offset, increasing viewpoint diversity. This off-axis placement exposes the back face of the real object to the camera. Crucially, the range of the camera's azimuth angle is constrained to ensure the object's reflection remains fully contained within the mirror's boundaries from the camera's perspective. Finally, the camera's optical axis is oriented to point toward the mirror's origin, with its elevation angle adjusted to ensure both the grounded primary objects and their reflections are optimally framed within the field of view.

\textbf{Depth map.}
The extracted depth map explicitly encodes the layered distance structure characteristic of a mirror scene: the real object is closest to the camera, the mirror surface lies at an intermediate depth, and the reflected object appears furthest behind the mirror. This distinct depth layering provides a clear spatial layout of the scene, which serves as a strong structural guide for the generation model.

\textbf{Segmentation map.}
Similarly, a per-region segmentation map is rendered, assigning a distinct color to each scene element: the real object, its reflection, mirror frame, floor, and surrounding walls. Each region is paired with a short descriptive text prompt, and the full mapping is exported as a JSON file.

\subsection{Conditioning-Guided Generation}
\label{sec:conditioning_guided_generation}

We adopt FLUX.1-dev~\cite{flux2024} as the backbone generative model due to its strong text-image alignment and supports high-resolution synthesis, making it well-suited for scenes with complex spatial relationships such as mirror reflections.

To translate our extracted 3D spatial priors into the final 2D image, we employ a zero-shot depth-conditioning strategy as our primary generation approach. Specifically, we use the pretrained OminiControl framework~\cite{tan2025ominicontrol}, which injects our rendered depth map as a lightweight plug-in adapter into FLUX.1-dev. Fig.~\ref{fig:conditioning-strategies} presents photorealistic images and their corresponding depth maps generated by this proposed method.

While our primary pipeline uses zero-shot depth conditioning due to its superior performance in maintaining geometric accuracy, we also investigated alternative ways. These included segmentation-conditioned generation (through the Seg2Any framework~\cite{li2025seg2any}) and a custom depth-conditioned Low-Rank Adaptation (LoRA) fine-tuned specifically on SynMirror dataset~\cite{dhiman2025reflecting}. Fig.~\ref{fig:conditioning-strategies} also presents representative segmentation maps and the corresponding final images generated by the segmentation-conditioned approach. 
\section{Proposed Benchmark Suite}

\begin{figure*}[tp]
    \centering
    \includegraphics[width=0.95\textwidth]{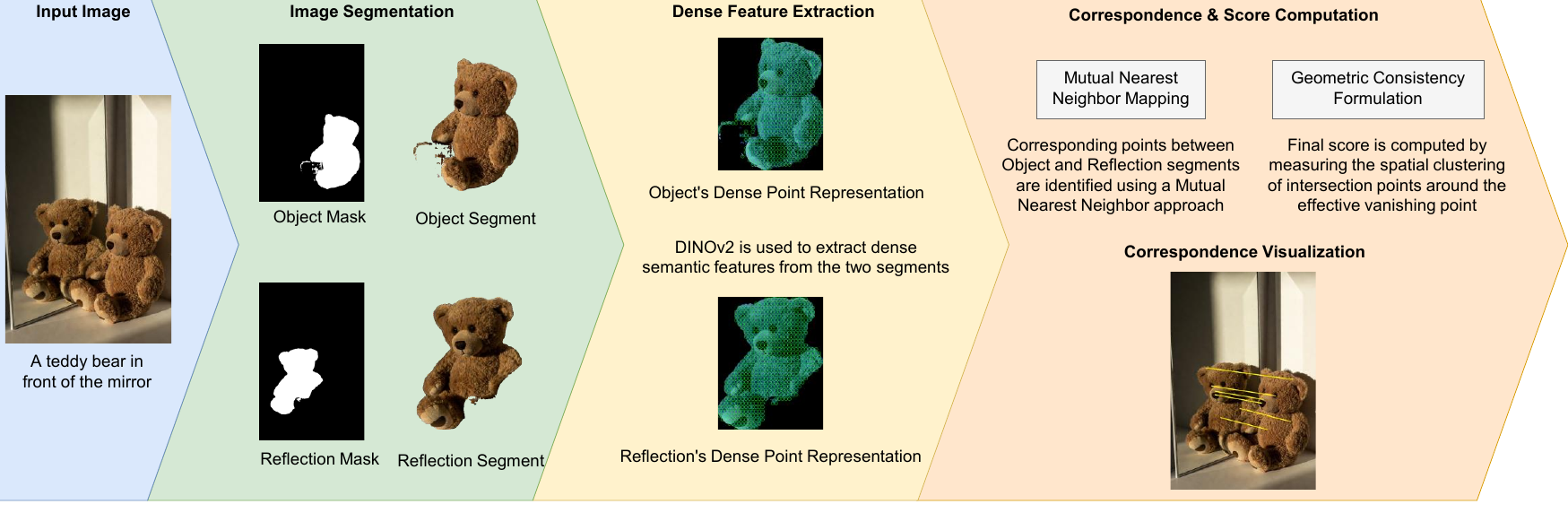}
    \vspace{-3mm}
    \caption{Pipeline of the proposed Mirror Consistency Score (MCS). The generated image is first segmented to isolate the primary objects and their reflections. Dense feature matching is then applied to establish corresponding keypoints, which are used to mathematically evaluate projective geometric consistency.}
    \label{fig:mcs}
    \vspace{-3mm}
\end{figure*}

\subsection{Dataset}

In this work, we introduce a new benchmark dataset, namely \textbf{Mirr}or \textbf{O}bject \textbf{B}enchmark (MirrOB). Instead of hand-picking random objects, our dataset is based on 80 standard object categories of the MS COCO dataset~\cite{lin2014microsoft} to ensure a wide variety of popular items.

Since generating mirror reflections becomes harder when there are more objects in the scene, we design our benchmark to include three levels of difficulty: single-object, two-object, and three-object scenes. To scale the prompt generation while maintaining natural language diversity, we take advantage of ChatGPT to generate textual prompts following a structured template: \textit{"a/an \{object\_1\}, ..., a/an \{object\_n\} in front of the mirror, \{scene\_description\}."} The dataset is divided into the following three categories:
\begin{itemize}
    \item \textbf{Single-Object Prompts:} For each of the 80 MS COCO object categories, we generate two distinct prompts. The first prompt features a simple scene description, while the second introduces a more complex, context-heavy background. This results in 160 single-object prompts.
    \item \textbf{Two-Object Prompts:} To test how well the models handle the correct spacing and angles for multiple items, we randomly sample pairs of objects from the 80 COCO categories and generate 100 unique two-object prompts.
    \item \textbf{Three-Object Prompts:} For the highest level of difficulty, we randomly sample triplets from the COCO categories and generate 100 unique three-object prompts.
\end{itemize}

In total, our benchmark comprises 360 structured prompts. This dataset allows us to measure how different generative models handle perspective, object details, and multi-object occlusion in mirror reflections.

\subsection{Mirror Consistency Score Metric}

\subsubsection{Motivation}

To quantitatively evaluate the geometric and physical correctness of mirror reflections without requiring human evaluation, we propose a novel, fully automated evaluation pipeline. Our metric leverages state-of-the-art models to establish dense semantic correspondences between a real object and its reflection, followed by a geometric analysis based on the principles of projective geometry. The pipeline of this metric is illustrated in Fig.~\ref{fig:mcs}.

\subsubsection{Object and Reflection Segmentation}
\label{sec:object_segment}
The first stage isolates the real object and its corresponding mirror reflection from the background. We employ Grounding DINO~\cite{liu2024grounding} for open-vocabulary detection and SAM~\cite{kirillov2023segment} for zero-shot segmentation.

The combined model outputs bounding boxes and precise boolean masks for the detected instances. We then apply the masks to the original image via element-wise multiplication, and extract tightly cropped regions of interest (ROIs).

\subsubsection{Dense Feature Extraction and Masking}
\label{sec:feature_extraction}
We utilize DINOv2~\cite{oquab2023dinov2}, which is well-known for its robust part-level feature representations. The cropped ROIs are resized and passed through the DINOv2 backbone. We extract the patch-level tokens, yielding a dense grid of $D$-dimensional feature descriptors. Because DINOv2 operates on a rigid patch grid, several extracted features inevitably fall on the suppressed background regions of our rectangular ROIs. We project the 1D sequence of patch tokens back into a 2D spatial grid and filter the keypoints using the alpha channel of the segmentation masks generated in the previous stage. This ensures that only descriptors strictly belonging to the object's surface are retained. Finally, the retained feature vectors are $L_2$-normalized to project them onto a unit hypersphere, which facilitates similarity computation.

\subsubsection{Bijective Feature Matching}
To find corresponding points between the real object and its reflection, we compute the cosine similarity between the sets of $L_2$-normalized descriptors. Let $F_r \in \mathbb{R}^{N \times D}$ and $F_m \in \mathbb{R}^{M \times D}$ be the filtered feature sets for the real object and its mirror reflection, respectively. The similarity matrix $S \in \mathbb{R}^{N \times M}$ is computed as: $S = F_r F_m^\top$.

To eliminate ambiguous or many-to-one matches (for example, multiple points on the object mapping to a single point on the reflection), we enforce a strict bijective mapping using a Mutual Nearest Neighbor (MNN) constraint. A match between descriptor $i$ in $F_r$ and descriptor $j$ in $F_m$ is only accepted if $i$ is the absolute nearest neighbor of $j$, and $j$ is the absolute nearest neighbor of $i$, satisfying a predefined similarity threshold $\tau$.

To isolate the most geometrically reliable connections, we sort the surviving mutually verified matches by their cosine distance and retain the top $K$ most confident pairs for each object in the image ($K$ is a configurable parameter).

\subsubsection{Geometric Consistency Formulation}
The main idea of our Mirror Consistency Score (MCS) is to measure the physical accuracy of the matched points. In 3D space, lines connecting points on a real object to their exact reflections on a flat mirror are strictly parallel. Under perspective projection in a 2D image, these parallel lines must converge at a single vanishing point. For this analysis, we explicitly assume the camera is positioned to avoid special geometric configurations, such as the camera's viewing direction being perfectly perpendicular to the mirror normal, that would prevent a finite vanishing point from forming.

First, we map the local coordinates of the matched keypoints back to the global image space. For each matched pair, we compute the normalized line equation $ax+by+c=0$. Because we process features on a per-object basis, an image containing $N$ valid objects generates a total of $N \times K$ geometric lines. Ideally, all lines intersect at a single coordinate. In practice, localization errors and optical distortions cause the intersections to form a distribution. We compute the pairwise intersections for all lines using Cramer’s rule. For lines $l_u$ and $l_v$, the intersection point $P_{u,v}=(x,y)$ is defined if the determinant of their coefficients is non-zero. Let $\mathcal{I}$ be the set of all valid intersection points. We calculate the geometric centroid $\mu$ (the effective vanishing point) as the mean of these intersections:

\begin{equation}
    \mu=\frac{1}{|\mathcal{I}|}\sum_{P \in \mathcal{I}}P.
\end{equation}

\begin{figure}[t]
    \centering
    \includegraphics[width=\linewidth]{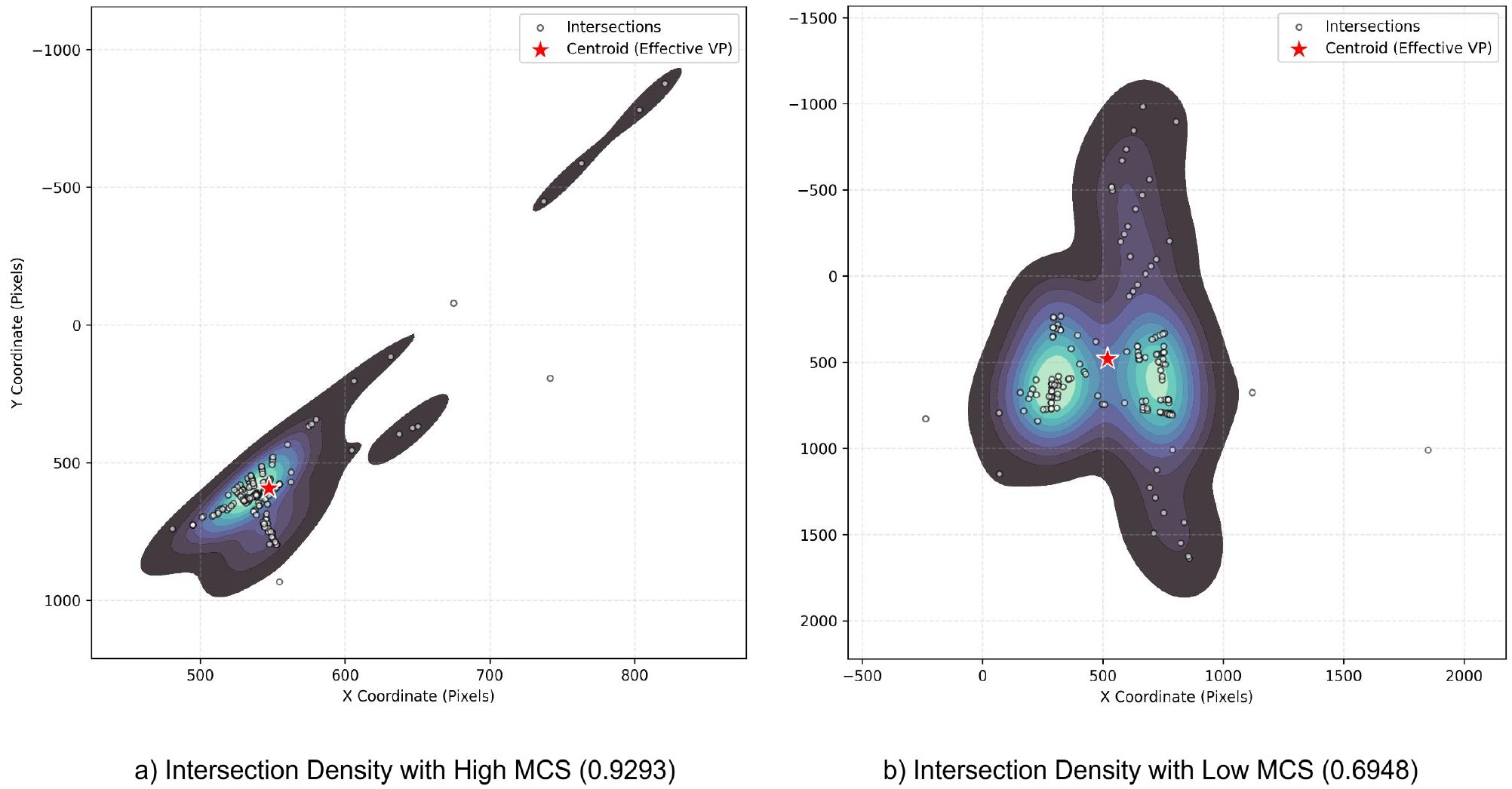}
    \vspace{-5mm}
    \caption{Visualization of the Mirror Consistency Score (MCS). (Left) A physically correct reflection yields tightly clustered intersection points around the vanishing point, resulting in a high MCS. (Right) Perspective mismatches cause the intersection points to scatter, which is penalized with a low score.}
    \label{fig:mcs_visualization}
    \vspace{-8mm}
\end{figure}

To quantify the consistency of the mirror reflection, we measure the tightness of this intersection cluster. To ensure the metric remains stable and invariant to the varying number of matched lines (which causes the number of pairwise intersections to grow quadratically in multi-object scenes), we compute the average of the Euclidean distances from each intersection point to the centroid. The final score is calculated using an exponential decay function to bound the metric between 0 and 1:

\begin{equation}
    MCS=\exp\left(-\frac{1}{\lambda |\mathcal{I}|}\sum_{P \in \mathcal{I}}\|P-\mu\|_2\right),
\end{equation}
where $\lambda$ is a normalization scaling factor. MCS approaching 1.0 indicates perfect projective consistency, while lower scores indicate structural warping, incorrect semantic generation, or physical impossibilities in the mirror reflection.

\subsubsection{Visualization}
To clearly show how our proposed metric works, Fig.~\ref{fig:mcs_visualization} displays a side-by-side comparison of two contour maps. On the left, an accurate image will result in a tight cluster of intersection points around the effective vanishing point, shown as a sharp peak, which results in a high MCS. On the right, when a model fails to follow mirror physics, the reflection angles are wrong, causing the intersection points to scatter randomly. Our formula penalizes this wide spread, leading to a low score.
\section{Experiments and Results}

\begin{figure*}[tp]
    \centering
    \includegraphics[width=\textwidth]{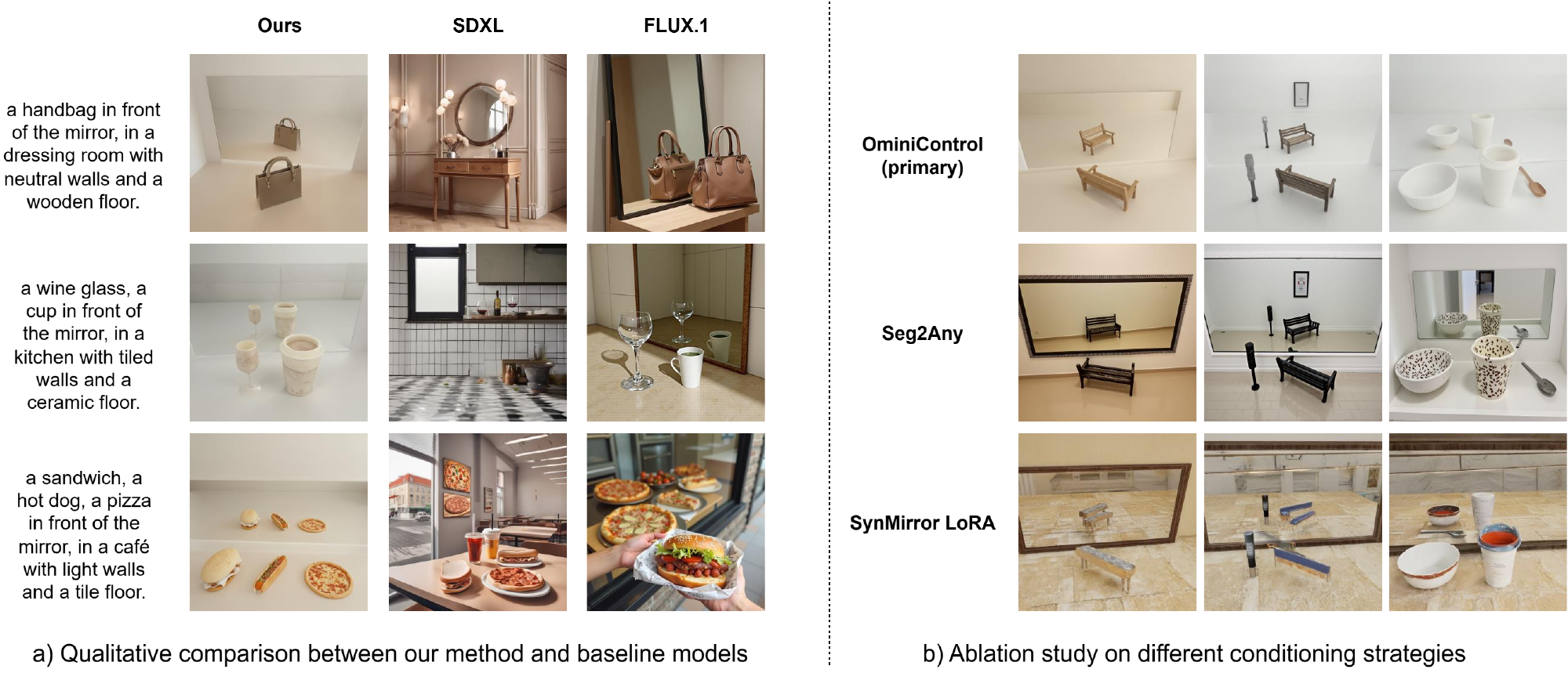}
    \vspace{-5mm}
    \caption{Qualitative comparison of synthesized mirror reflections across varying scene complexities (1 to 3 objects). (a) Unconditioned baselines (SDXL and FLUX.1-dev) frequently fail to produce physically accurate reflections, whereas our physics-aware method consistently generates geometrically correct results. (b) Ablation study comparing alternative spatial conditioning strategies, demonstrating how segmentation priors (Seg2Any) and custom fine-tuning (SynMirror LoRA) can introduce visual artifacts or perspective errors compared to our primary method.}
    \label{fig:qualitative_results}
\end{figure*}
\begin{table*}
\caption{Quantitative results on Mirror Consistency Score, CLIP Score, and image quality metrics. Best results are highlighted in bold, second-best results are underlined. Higher scores ($\uparrow$) indicate better performance.}
\centering
\small
\adjustbox{width=0.9\textwidth}{\begin{tblr}{
  cells = {c},
  cell{1}{1}  = {r=2}{},               
  cell{1}{2}  = {c=4}{},               
  cell{1}{6}  = {r=2}{},               
  cell{1}{7}  = {c=3}{},               
  vline{2}    = {1-8}{solid},          
  vline{6}    = {1-8}{solid},          
  vline{7}    = {1-8}{solid},          
  hline{1,7}  = {-}{0.08em},           
  hline{3}    = {-}{},                 
}
\textbf{Method}                    & \textbf{Mirror Consistency Score}          &           &           &                    & \textbf{CLIP Score} $\uparrow$ & \textbf{Image Quality}      &                   &                   \\
                         & \textbf{1 Obj} $\uparrow$ & \textbf{2 Obj} $\uparrow$ & \textbf{3 Obj} $\uparrow$ & \textbf{Overall} $\uparrow$ &                       & \textbf{CLIP-IQA} $\uparrow$ & \textbf{MANIQA} $\uparrow$ & \textbf{MUSIQ} $\uparrow$  \\
SDXL                     & 0.480            & 0.580            & 0.548            & 0.527              & 33.284    & 0.6499  & \textbf{0.5114}   & \textbf{73.4996}  \\
SD3.5-Large              & \underline{0.524}            & 0.679            & 0.693            & \underline{0.614}              & \textbf{34.1322}    & \underline{0.6333}  & 0.4163   & 70.3584  \\
FLUX.1-dev               & 0.476            & \underline{0.689}            & \underline{0.670}            & 0.589              & \underline{33.361}        & 0.6051              & 0.4733            & 70.5174           \\
\textbf{PhysMirror (Ours)}                     & \textbf{0.727}& \textbf{0.779}   & \textbf{0.744}   & \textbf{0.746}     & 28.914                & \textbf{0.6589}     & \underline{0.4929}& \underline{70.7800}\\
\end{tblr}}

\label{tab:quantitative}
\end{table*}

\subsection{Implementation Details}
\label{sec:implementation_details}
\textbf{Hardware Setup.}
All experiments, including image generation, metric evaluation, and fine-tuning procedure, were conducted on a single NVIDIA A100 GPU.

\textbf{3D Scene Construction.}
For the 3D mesh generation stage, we employed TRELLIS~\cite{xiang2025structured} as our primary text-to-3D model due to its strong shape fidelity. The implementation also supports Shap-E~\cite{jun2023shap} as a lightweight alternative. To construct the simulated 3D scene and render spatial conditioning maps (depth and segmentation), we used PyTorch3D library~\cite{ravi2020accelerating}.

\textbf{Primary Generation Pipeline and Evaluation.}
For our primary geometry-conditioned image synthesis, we used the pretrained OminiControl framework~\cite{tan2025ominicontrol}, applying its official depth LoRA weights to the FLUX.1-dev backbone. For the automated evaluation of these images, our proposed Mirror Consistency Score (MCS) metric utilized SAM 2.1 large~\cite{kirillov2023segment} for zero-shot semantic segmentation and DINOv2 large~\cite{oquab2023dinov2} for dense feature extraction. We computed the MCS using a mutual nearest neighbor matching threshold of $\tau=0.5$ and evaluated the top $K=10$ keypoint pairs per object.

\textbf{Ablation Baselines.}
To support our ablation studies (detailed in Section~\ref{sec:ablation}), we implemented two alternative conditioning strategies. First, for segmentation-based conditioning, we employed the Seg2Any framework~\cite{li2025seg2any} using weights pretrained on the SACap-1M dataset. Second, for a custom depth-conditioned alternative, we finetuned a LoRA module on FLUX.1-Depth-dev pipeline from the Hugging Face Diffusers library. We selected 3,000 high-quality images from the SynMirror dataset~\cite{dhiman2025reflecting}, extracting their corresponding depth maps using Depth Anything V2~\cite{yang2024depth}. The LoRA was trained to minimize the denoising loss for 5 epochs with a learning rate of $1 \times 10^{-4}$, a batch size of 1, 4 gradient accumulation steps, and a rank of 64.

\subsection{Experimental Setup}
\textbf{Baselines.} We evaluated the proposed PhysMirror on the MirrOB dataset, comparing it against three state-of-the-art text-to-image foundation models evaluated out-of-the-box: FLUX.1-dev, Stable Diffusion XL (SDXL), and Stable Diffusion 3.5 Large (SD3.5-Large).

\textbf{Metrics.} Performance was assessed across three primary dimensions: geometric correctness, semantic alignment, and perceptual image quality. To measure geometric correctness and physical realism, we applied our proposed Mirror Consistency Score (MCS). To examine the model's text-to-image alignment, we computed the CLIP Score~\cite{radford2021learning}. Finally, we evaluated overall perceptual image quality using standard reference-free metrics, specifically CLIP-IQA~\cite{wang2023exploring}, MANIQA~\cite{yang2022maniqa}, and MUSIQ~\cite{ke2021musiq}.

\subsection{Quantitative Results}
As summarized in Table~\ref{tab:quantitative}, we evaluate the unconditioned baseline models against our proposed physics-aware pipeline across varying levels of scene complexity (1, 2, and 3 objects)

\textbf{Geometric Correctness.} Our proposed method consistently outperforms the baseline models in geometric accuracy across all difficulty levels. It achieves the highest overall Mirror Consistency Score of 0.746 and demonstrates robustness in complex multi-object scenarios, scoring 0.779 for two-object and 0.744 for three-object scenes. In contrast, baseline models struggle to process projective symmetry, leading to significantly lower overall scores of 0.527 for SDXL and 0.589 for FLUX.1-dev.

\textbf{Semantic Alignment.} Conditioned generative models often suffer from degraded text-to-image alignment. This trade-off is reflected in our CLIP Score evaluations. The baseline models achieve the highest semantic alignment scores. However, our proposed method maintains a competitive CLIP score of 28.914, which shows that explicit 3D priors can enforce physical laws without catastrophically degrading text-image consistency.

\textbf{Image Quality.} Despite the added geometric constraints, our method achieves the highest CLIP-IQA score (0.659). This indicates that our method produces more perceptually natural images. SDXL scores slightly higher on MANIQA and MUSIQ, suggesting that its outputs are smoother at the pixel level.

\subsection{Qualitative Results}
To visually validate our quantitative findings, we present a qualitative comparison of the synthesized images. Fig.~\ref{fig:qualitative_results} (a) compares our primary method against the unconditioned baselines.

\textbf{Baseline Limitations.} SDXL successfully generates semantic objects but consistently fails to position them in front of a mirror or render reflections. FLUX.1-dev demonstrates better structural understanding. At first glance, it appears to synthesize appropriate reflection scenes for one-object and two-object prompts. However, closer inspection reveals geometric inconsistencies between the primary objects and their mirrored counterparts. Furthermore, FLUX.1-dev severely struggles to maintain spatial coherence when processing complex three-object scenarios.

\textbf{Impact of Physics-Aware Conditioning.} In contrast, our pipeline consistently generates geometrically sound reflections. By anchoring the generation to an explicit 3D scene, our method enforces strict perspective projection. As shown in Fig.~\ref{fig:qualitative_results} (a), our method produces the most physically accurate and photorealistic images across all difficulty levels without sacrificing semantic fidelity.

\subsection{Ablation Study}
\label{sec:ablation}

\begin{table*}
\caption{Quantitative results of ablation study. Best results are highlighted in bold, second-best results are underlined.}
\centering
\small
\adjustbox{width=\textwidth}{%
\begin{tblr}{
  cells = {c},
  cell{1}{1}  = {c=2}{},   
  cell{1}{3}  = {c=4}{},   
  cell{1}{7}  = {r=2}{},   
  cell{1}{8}  = {c=3}{},   
  cell{1}{11} = {r=2}{},   
  vline{3}    = {1-7}{solid},
  vline{7}    = {1-7}{solid},
  vline{8}    = {1-7}{solid},
  vline{11}   = {1-7}{solid},
  hline{1,7}  = {-}{0.08em},
  hline{3}    = {-}{},
}
\textbf{Configuration}
& 
& \textbf{Mirror Consistency Score}
& & & &
\textbf{CLIP Score} $\uparrow$
& \textbf{Image Quality}
& & &
\textbf{Time (s)} $\downarrow$ \\
\textbf{Strategy}
& \textbf{Text-to-3D}
& \textbf{1 Obj} $\uparrow$
& \textbf{2 Obj} $\uparrow$
& \textbf{3 Obj} $\uparrow$
& \textbf{Overall} $\uparrow$
&
& \textbf{CLIP-IQA} $\uparrow$
& \textbf{MANIQA} $\uparrow$
& \textbf{MUSIQ} $\uparrow$
& \\
Seg2Any
& TRELLIS
& 0.540
& 0.719
& \underline{0.700}
& 0.634
& \textbf{31.924}
& 0.6115
& 0.4851
& \textbf{74.1957}
& - \\
SynMirror LoRA
& TRELLIS
& \textbf{0.729}
& \underline{0.732}
& 0.637
& \underline{0.704}
& \underline{30.307}
& 0.5344
& 0.4392
& 69.9332
& - \\
OminiControl
& TRELLIS
& \underline{0.727}
& \textbf{0.779}
& \textbf{0.744}
& \textbf{0.746}
& 28.914
& \textbf{0.6589}
& \textbf{0.4929}
& \underline{70.7800}
& \underline{121.943} \\
OminiControl
& Shap-E            
& 0.670          
& 0.625          
& 0.596          
& 0.637          
& 28.003          
& \underline{0.6569}          
& \underline{0.4888}          
& 67.7559        
& \textbf{96.438} \\
\end{tblr}}
\label{tab:ablation}
\vspace{-5mm}
\end{table*}

To evaluate the impact of different spatial priors, we compare our primary zero-shot depth-based approach (OminiControl) against a zero-shot segmentation-conditioned method (Seg2Any) and a custom fine-tuned depth adapter (SynMirror LoRA). Quantitative results are summarized in Table~\ref{tab:ablation}, while qualitative results are illustrated in Fig.~\ref{fig:qualitative_results} (b).

\textbf{Depth vs. Segmentation Priors.}
Although the Seg2Any framework achieves the highest CLIP Score (31.924), it lacks explicit geometric reasoning. Because segmentation maps do not encode depth relationships, the generated images often appear spatially inconsistent, with unnatural object placement. Moreover, Seg2Any frequently introduces visual artifacts in reflection regions. These artifacts disrupt dense feature matching in our Mirror Consistency Score metric, leading to wrong correspondence estimation between objects and their reflections. Consequently, Seg2Any achieves the lowest overall MCS (0.634).

\textbf{Zero-Shot vs. Fine-Tuned Depth.}
SynMirror LoRA demonstrates strong geometric consistency in simple single-object scenes (MCS: 0.729), indicating that dataset-specific fine-tuning can effectively encode basic reflection priors. However, its performance degrades significantly as the  complexity of the scene increases. In multi-object scenarios, the model fails to preserve object completeness, occasionally omitting objects or generating distorted shapes. In contrast, the zero-shot OminiControl framework maintains object integrity and spatial coherence even in complex three-object settings, achieving the highest overall MCS (0.746).

\textbf{Image Quality.} OminiControl achieves the highest CLIP-IQA (0.659) and MANIQA (0.493) scores among all conditioning variants, confirming that depth-based conditioning preserves perceptual quality most effectively. Seg2Any achieves the best MUSIQ score (74.196), likely due to its less constrained generation process producing smoother textures. SynMirror LoRA obtains the lowest IQA scores across all three metrics, consistent with the visual artifacts observed in its outputs.

\revise{\textbf{Text-to-3D Backbone Comparison.} To quantify the trade-offs between initial 3D mesh quality, final generation accuracy, and computational efficiency, we evaluated our pipeline using two distinct text-to-3D backbones. As detailed in Table \ref{tab:ablation}, TRELLIS consistently outperforms the lightweight Shap-E alternative across all spatial and perceptual metrics. TRELLIS achieves a significantly higher overall MCS (0.746 compared to 0.637) and higher image quality scores. However, this geometric precision comes at a computational cost: Shap-E reduces the pipeline's inference time to 96.4 seconds, making it a high-speed alternative to the 121.9 seconds required by TRELLIS.}



\section{Conclusion}
In this paper, we addressed the challenge of generating physically correct mirror reflections in modern text-to-image diffusion models. To this end, we proposed PhysMirror which is a novel and physics-aware pipeline. PhysMirror lifts text prompts into 3D meshes to construct a geometrically correct mirrored scene. By extracting precise spatial conditioning maps from this 3D environment, we successfully guided foundation models like FLUX.1-dev to respect projective physics. Furthermore, to address the absence of quantitative evaluation metrics in this domain, we introduced the Mirror Consistency Score (MCS). This fully automated metric leverages dense feature matching and vanishing point theory to evaluate the geometric correctness of reflections.

Extensive evaluations demonstrate that our proposed PhysMirror significantly improves reflection realism over unconditioned baselines. Notably, our primary approach using zero-shot depth-conditioned OminiControl framework maintains robust structural coherence even in complex, multi-object environments. Future work will explore extending this explicit 3D conditioning to non-planar reflective surfaces, complex lighting interactions, and temporal geometric consistency for video generation.

\section*{Acknowledgments}

\revise{This research was carried out as part of the joint research program between the University of Science, Vietnam National University Ho Chi Minh City (VNU-HCM), and the Center for AI Research (CAIR), VinUniversity.}

\bibliographystyle{IEEEtran}
\bibliography{IEEEabrv, reference}

\end{document}